# *New Directions in Text Classification Research: Maximizing The Performance of Sentiment Classification from Limited Data*

# Arah Baru Penelitian Klasifikasi Teks: Memaksimalkan Kinerja Klasifikasi Sentimen dari Data Terbatas

**Surya Agustian[1]\*, Muhammad Irfan Syah[2], Nurul Fatiara[3], Rahmad Abdillah[4]**

[1,2,3,4]Teknik Informatika, UIN Sultan Syarif Kasim Riau, Indonesia

E-Mail: [1]surya.agustian@uin-suska.ac.id, [2]mirfansyah1182@gmail.com, [3]12050122431@uin-suska.ac.id, [4]rahmad.abdillah@uin-suska.ac.id

*Corresponding Author: Surya Agustian*

**Abstract**

*The stakeholders' needs in sentiment analysis for various issues, whether positive or negative, are speed and accuracy. One new challenge in sentiment analysis tasks is the limited training data, which often leads to suboptimal machine learning models and poor performance on test data. This paper discusses the problem of text classification based on limited training data (300 to 600 samples) into three classes: positive, negative, and neutral. A benchmark dataset is provided for training and testing data on the issue of Kaesang Pangarep's appointment as Chairman of PSI. External data for aggregation and augmentation purposes are provided, consisting of two datasets: the topic of Covid Vaccination sentiment and an open topic. The official score used is the F1-score, which balances precision and recall among the three classes, positive, negative, and neutral. A baseline score is provided as a reference for researchers for unoptimized classification methods. The optimized score is provided as a reference for the target score to be achieved by any proposed method. Both scoring (baseline and optimized) use the SVM method, which is widely reported as the state-of-the-art in conventional machine learning methods. The F1-scores achieved by the baseline and optimized methods are 40.83% and 51.28%, respectively.*

**Abstrak**

Kebutuhan pihak terkait dalam menganalisis sentimen terhadap berbagai isu, apakah positif atau negatif, adalah kecepatan dan ketepatan. Beberapa tantangan baru dalam tugas analisis sentimen, salah satunya adalah terbatasnya data *training*, yang sering kali menyebabkan model pembelajaran mesin tidak optimal, menghasilkan kinerja yang rendah pada data uji. Paper ini membahas problem klasifikasi teks berdasarkan ketersediaan data *training* yang terbatas (300 sampai 600 sampel), ke dalam 3 kelas: positif, negatif dan netral. Benchmark dataset diberikan untuk data training dan testing dengan isu pengangkatan Kaesang Pangarep sebagai Ketua Umum PSI. Data eksternal untuk kebutuhan agregasi dan augmentasi disediakan, yang terdiri atas 2 set data, yaitu topik sentimen program Vaksinasi Covid dan topik terbuka (Open Topic). Official *Score* yang digunakan adalah F1-*score*, yang menghitung secara berimbang precision dan recall di antara ketiga kelas, positif, negatif dan netral. Baseline *score* diberikan untuk menjadi acuan bagi peneliti, bagi metode klasifikasi yang tidak dioptimasi. Sedangkan optimized *score* diberikan untuk menjadi acuan *score* yang perlu dikejar dari setiap metode yang diusulkan. Kedua scoring (baseline dan optimized) menggunakan metode SVM, yang banyak dilaporkan sebagai state-of-the-art dari metode machine learning konvensional. Hasil F1-*score* yang dicapai dari metode Baseline dan Optimized adalah 40.83% dan 51.28%.

Kata Kunci: klasifikasi sentimen, dataset terbatas, SVM, benchmark dataset

## 1. INTRODUCTION

Penggunaan platform media sosial melalui internet telah meluas dengan akses yang mudah ke berbagai belahan di dunia. Media sosial menjadi pusat penyebaran berbagai konten, topik dan isu dalam berbagai format dan bahasa. Teks dalam *tweet* dapat mendeskripsikan pengalaman pribadi, promosi/pemasaran produk, sampai kepada umpan balik dari suatu pembelian barang [1]. Masyarakat juga memanfaatkan media sosial untuk mencari referensi dan menentukan preferensi, mengenai kualitas suatu barang atau jasa, baik tidaknya layanan dari suatu perusahaan maupun lembaga pemerintah, isu-isu politik dan sosial, dan sebagainya. Media sosial juga sering kali dipakai untuk penyebaran berita dan *hoax*, propaganda, sampai kepada konten yang menyerang pribadi maupun golongan. Dalam bidang sosial dan politik seperti pemilu, postingan *tweet* masih dipakai oleh para *stake holder* dan yang berkepentingan [2], dalam mengukur tingkat kepuasan dan sentimen masyarakat terhadap isu-isu tertentu.

Untuk memenuhi kebutuhan pihak terkait dalam menganalisis sentimen terhadap berbagai isu, apakah positif atau negatif, tentu saja membutuhkan kecepatan dan ketepatan. Sebagai contoh dalam pemilihan pemimpin suatu negara, tim pemenangan akan mencoba menganalisa popularitas dan sentimen terhadap calon yang diusung, salah satunya dari media sosial. Dan tentu saja, hasil yang diharapkan harus diperoleh dengan cepat, tidak bisa menunggu lama. Oleh karena itu, pengukuran sentimen menggunakan mesin (komputer) menjadi andalan, walaupun dari segi ketepatan analisis masih perlu dipelajari lebih mendalam. Hal ini melahirkan berbagai penelitian dan *shared task* analisis sentimen dengan menggunakan bantuan komputer, yang banyak melibatkan metode pembelajaran mesin atau kecerdasan buatan.

Penelitian analisis sentimen lebih banyak berfokus pada studi kasus atau topik, hanya melakukan klasifikasi sentimen berdasarkan kelas, seperti sentimen positif, negatif dan netral [3], [4], atau hanya kelas positif dan negatif saja [5], [6]. Beberapa tantangan baru dalam tugas analisis sentimen, salah satunya adalah terbatasnya data *training*, yang sering kali menyebabkan model pembelajaran mesin tidak optimal, menghasilkan kinerja yang rendah pada data uji, bahkan tidak dapat menghasilkan prediksi yang akurat. Beberapa alasan mengapa sumber data training mungkin terbatas, bisa disebabkan oleh faktor biaya pengumpulan data, keterbatasan sumber daya untuk menilai dan mengevaluasi data, keterbatasan akses, atau kesulitan dalam mengumpulkan data yang relevan. Termasuk juga alasan kecepatan dalam penyediaan data training untuk pembelajaran mesin yang sudah diberikan label kelasnya. Suatu lembaga survey tidak akan menghabiskan waktu dan biaya untuk melakukan *labeling* pada data yang banyak, yang mencukupi untuk menghasilkan model machine learning yang handal, dari setiap kasus atau isu yang akan dianalisis sentimennya. Apabila dilakukan, bukan tidak mungkin kastemer yang merupakan tokoh publik atau tim pemenangan kampanye akan berpindah kepada lembaga survey lain yang dapat menghasilkan analisis lebih cepat.

Dalam penelitian *shared task* ini, keterbatasan jumlah data training ditetapkan sebagai tantangan, yang muncul atas kebutuhan *real* para pelanggan terkait waktu, yaitu untuk mengetahui sentimen terhadap suatu isu dengan cepat. Sedangkan bagi lembaga survey yang melakukan analisis sentimen, mereka akan berusaha menekan biaya untuk membayar *anotator*, sehingga keuntungan jasa menganalisis sentimen menjadi lebih besar. Oleh karena itu, komputer harus dapat mengklasifikasi sentimen berdasarkan model hasil training, dari data yang mungkin hanya sempat diberi label (positif, negatif atau netral) secara manual dalam jumlah yang minimal oleh manusia (*anotator*). Padahal, untuk memberikan label dengan jumlah data mencukupi untuk melatih mesin dari suatu metode ML, akan membutuhkan biaya, waktu dan sumber daya yang lebih banyak.

Paper ini membahas problem klasifikasi teks berdasarkan ketersediaan data *training* yang terbatas (300 sampai 600 sampel), ke dalam 3 kelas: positif, negatif dan netral, sebagaimana tantangan dalam klasifikasi *hate speech* pada *shared task* HASOC 2023 [7]. Penelitian ini juga dijalankan dalam bentuk *shared task*, untuk menarik minat beberapa author/peneliti untuk berpartisipasi mengembangkan metode masing-masing. Salah satu kontribusi dari penelitian ini adalah suatu set data *tweet* yang dapat dijadikan *benchmark* dalam klasifikasi sentimen. Studi kasus yang digunakan adalah isu pengangkatan Kaesang Pangarep, yang merupakan putra presiden Jokowi, sebagai ketua umum PSI. Isu ini menjadi viral dan menuai pro dan kontra (sentimen) di kalangan masyarakat, karena proses pengangkatan yang tidak melalui mekanisme yang umum dan wajar dalam sebuah partai politik. Ketidakwajaran itu adalah saat yang bersangkutan baru saja direkrut menjadi anggota partai, dan kemudian langsung diangkat menjadi ketua umum tanpa melalui proses kaderisasi partai maupun pemilihan umum partai.

Paper ini diorganisasikan dalam susunan sebagai berikut. Bagian pertama adalah pendahuluan, yang menerangkan latar belakang penelitian dan pelaksanaan *shared task* ini, serta beberapa review atas penelitian yang terkait. Bagian selanjutnya menerangkan material penelitian yang digunakan dalam *shared task* ini, yaitu mengenai *dataset*, kualitas data, dan metode baseline yang diterapkan sebagai acuan evaluasi bagi para peneliti lain. Bagian ketiga menerangkan tentang hasil penerapan metode baseline dan *score* evaluasi yang diperoleh. Kesimpulan dan saran-saran bagi para peneliti yang tertarik untuk mengembangkan metode klasifikasi untuk *problem* yang diangkat dalam *shared task* ini, disampaikan dalam bagian terakhir.

## 2. MATERIAL DAN METODE

Analisis sentimen adalah penerapan ilmu komputer di dalam bidang linguistik (bahasa) yang bertujuan untuk memahami dan menganalisis opini, perasaan, dan sentimen yang terkandung dalam teks. *Use case* dari analisis sentimen dapat diterapkan pada berbagai macam data teks, mulai dari ulasan produk, tinjauan film, dukungan untuk tokoh politik, isu yang menerpa kehidupan artis, sampai kepada tanggapan masyarakat terhadap program dan kebijakan pemerintah. Teks-teks tersebut dapat digali dari media sosial, forum komunikasi, umpan balik aplikasi telepon pintar, sampai komentar-komentar di situs web. Salah satu tujuan dari sentimen analisis adalah untuk mengklasifikasikan teks menjadi kategori-kategori seperti positif, negatif, atau netral.

Tantangan utama dalam sentimen analisis adalah keterbatasan sumber data training. Sumber data training yang kecil dapat mengakibatkan model yang dihasilkan memiliki kinerja yang rendah atau bahkan tidak dapat menghasilkan hasil yang akurat. Beberapa alasan mengapa sumber data training mungkin terbatas termasuk biaya pengumpulan data, keterbatasan akses, kesulitan dalam mengumpulkan data yang relevan atau peniliaian yang objektif, keterbatasan sumber daya ahli, dan sebagainya.

Beberapa pendekatan dan langkah-langkah optimasi yang dapat digunakan untuk mengatasi masalah sumber data latih yang kecil dalam sentimen analisis termasuk:

a. *Transfer Learning*: memanfaatkan model yang telah dilatih sebelumnya pada *task-task* terkait atau menggunakan *pre-trained language models* seperti BERT [8], GPT [9], atau *word embeddings* yang telah dilatih pada *dataset* yang sangat besar.
b. *Data Augmentation*: menggunakan teknik augmentasi data seperti *sinonim substitution*, penggabungan kalimat, atau penambahan *noise* untuk memperluas *dataset training*.
c. *Data Aggregation*: memanfaatkan data eksternal dari topik yang berbeda untuk menambah data latih yang terbatas.
d. *Domain Adaptation*: mengadaptasi model yang telah dilatih pada domain yang berbeda agar dapat bekerja lebih baik pada *dataset* yang lebih kecil atau domain yang spesifik.

Sedangkan dari segi proses klasifikasi, terdapat banyak opsi langkah optimasi yang dapat dilakukan. Opsi pertama misalnya analisis dan pemilihan fitur klasifikasi (fitur kata/token, word embeddings, fitur kamus leksikon, fitur emoji, fitur tanda baca yang mengandung sentimen) [10]. Opsi selanjutnya dapat berupa rekayasa fitur (*feature engineering*) yang menggabungkan beberapa jenis fitur yang berbeda proses pembentukan dan cara perhitungannya, seperti pada [11], [12]. Optimasi metode machine learning termasuk ke dalamnya hyper-parameter tuning [13], analisis dan optimasi dataset hasil penggelembungan data sample akibat proses agregasi maupun augmentasi [14], dan sebagainya, juga dapat dilakukan untuk meningkatkan performa klasifikasi yang diinginkan.

Selain opsi-opsi yang telah banyak diperkenalkan oleh para peneliti sebelumnya, kita juga dapat melakukan pendekatan empiris, yang berdasarkan pengalaman dan coba-coba berdasarkan hipotesa. Tentu saja hipotesa yang dilakukan harus berdasarkan pengalaman dan *rationale* (alasan atau justifikasi) yang tepat sesuai dengan kasus yang dihadapi. Kemudian hipotesa tersebut dikembangkan dalam sebuah pendekatan algoritma untuk memecahkan problem klasifikasi yang diberikan.

### 2.1. Benchmark Dataset

Sebagai gambaran bagi para peneliti yang berpartisipasi dalam *shared task* ini, kami sebagai *organizer* menyediakan *dataset* untuk proses training metode ML, dan sejumlah data uji sebagai *benchmark*, sehingga semua metode dapat diukur dan dibandingkan performanya. *Dataset* yang tersedia sebagai data latih dan data uji, dan distribusinya, dapat dilihat pada Tabel 1 di bawah ini, yang dapat didownload pada situs Github[1].

**Tabel 1.** *Dataset* **penelitian**

| No | Dataset | Penggunaan | Jumlah sampel data | Distribusi kelas | | |
|---|---|---|---|---|---|---|
| | | | | Positif | Netral | Negatif |
| 1. | *Dataset* Kaesang v1 | *Training* | 300 | 100 | 100 | 100 |
| 2 | *Dataset* Kaesang v2 | *Training* | 300 | 100 | 100 | 100 |
| 3 | *Dataset* Program Vaksin Covid | *Training* | 8000 | 463 | 6664 | 873 |
| 4 | *Dataset Open Topic* | *Training* | 7569 | 1505 | 3408 | 2656 |
| 5 | *Dataset* Kaesang | *Testing* | 924 | | | |

#### 2.1.1 Dataset Kaesang

Sesuai dengan yang dibahas di bagian pendahuluan, bahwa sebagai studi kasus dalam *shared task* ini adalah mengenai pengangkatan Kaesang sebagai Ketua Umum PSI. Data teks sejumlah 2309 sampel, bersumber dari media sosial *Twitter*, yang dikumpulkan melalui proses *scrapping* mulai dari tanggal 25 September – 3 Oktober 2023, dengan menggunakan kata kunci "Kaesang PSI". Pada tahap awal, data akan diberikan label oleh beberapa orang *anotator* secara crowdsourcing. Untuk memudahkan pembagian data kepada para *anotator*, maka *tweet* yang diambil digenapkan sebanyak 2000 sampel. Sesuai dengan rancangan *shared task* ini, yaitu untuk pengembangan sistem dan model klasifikasi menggunakan *dataset training* yang terbatas, dari 2000 *tweet* di atas, sebagian kecil akan dijadikan data *training*, dan sisanya untuk pengujian (*testing*). Data *training* beserta label kelasnya akan diberikan kepada para peneliti yang berpartisipasi dalam *shared task*, sedangkan data *testing* tidak diberikan label kelasnya. Proses evaluasi akan dilakukan secara terpusat dalam suatu sistem *leaderboard* (papan *score*), dan *scoring* dilakukan secara otomatis.

---

[1] https://github.com/s4gustian/Small_DataSet_Sentiment_Classification

### Crowd Sourcing untuk Anotasi Data

Dari 2000 *tweet* yang telah dikumpulkan, dilakukan pemberian label kelas positif, negatif dan netral dengan cara *crowdsourcing* yang melibatkan 14 orang *anotator*. *Crowdsourcing* merupakan salah satu cara yang populer dalam membangun *dataset*, karena dapat mengurangi biaya khususnya pada bahasa-bahasa yang mayoritas di dunia [15].

Tim *anotator* untuk memberikan label pada isu sentimen ini, dipilih dari kalangan mahasiswa, dengan syarat bahwa mereka adalah penutur asli (Bahasa Indonesia) dan diberikan *briefing* yang cukup termasuk untuk bersikap objektif terhadap isu yang dibicarakan di dalam *tweet*. Setiap *tweet* asli (tanpa *text preprocessing*) akan dilabel oleh 4 orang *anotator*, dan label final ditentukan berdasarkan suara terbanyak (*majority voting*) [16].

### Majority Voting

Dalam memilih label final, akan terdapat satu kondisi suatu *tweet* tidak memiliki kelas yang mayoritas, yaitu ada 2 kelas dipilih oleh masing-masing 2 *anotator*. Untuk *tweet* seperti ini, di mana tidak ada kelas yang dominan, maka akan dihapus dari *dataset* sebagaimana di dalam penelitian Nabil dkk [16]. Sedangkan untuk komposisi *voting* lainnya, tidak ada masalah dalam penerapan *majority vote*. Untuk data dengan label kelas lebih dari 2, maka berapa pun jumlah *anotator* untuk setiap *tweet*, tetap akan mengalami permasalahan seperti tersebut. Sedangkan untuk klasifikasi teks dengan 2 kelas (misalnya positif dan negatif saja), maka 3 *anotator* sudah cukup sebagai angka minimal.

Penelitian ini menggunakan 3 kelas (positif, negatif dan netral), dan menggunakan 4 orang *anotator* agar dapat menghasilkan label data yang lebih berkualitas. Sebenarnya, 3 orang *anotator* sudah cukup untuk membangun *dataset* menggunakan suara terbanyak. Namun agar label hasil penilaian lebih kuat dan akurat, maka dilakukan penentuan label oleh 4 *anotator*, yang dapat diilustrasikan dalam Tabel 2 berikut ini. Pada contoh kalimat nomor 4, di mana terdapat penilaian sentimen yang sama banyak (2 positif dan 2 netral), maka *tweet* dihapus dari dataset.

Tabel 2. Ilustrasi penentuan label *tweet* secara *crowdsourcing* dari 4 orang *anotator*

| No | Kalimat *tweet* | Anotatator | | | | Voted label |
|---|---|---|---|---|---|---|
| | | A1 | A2 | A3 | A4 | |
| 1 | @HusinShihab @kaesangp @psi_id Selamat Mas Kaesang. Pesan saya hanya satu, jangan ajak si husin bokep ini gabung ke partaimu ya mas. | **positif** | netral | negatif | **positif** | **positif** |
| 2 | @ffarliani Lucuan Kaesang ngomong didepan kader PSI, berasa lagi stand up komedi ðŸ˜… | positif | **netral** | negatif | **netral** | **netral** |
| 3 | @bobby_risakotta @Cerdas007Cermat @jokowi @ganjarpranowo Cuman tunggu waktu saja Kaesang tarik kembali PSI dukung GP. Ngak juga gpp sih? Ngak ngaru.ðŸ˜Š | **negatif** | **negatif** | netral | **negatif** | **negatif** |
| 4 | Saat representative @TrisaktiMedia #GenZ tidak ingin hanya sorotan komoditas dan #Tvone meminta expresi ungkapan kaderisasi dari arahan @psi_id pembina, disuguhkan sosok @kaesang_id kurang tepat dalam spontan arahan pembinaan @psi | netral | netral | **positif** | **positif** | [dihapus] |

Setelah label final dipilih sesuai dengan kriteria di atas, diperoleh data *tweet* sebanyak 1892 sampel. Kemudian, dilakukan *fine grain processing*, untuk menghasilkan data *tweet* yang unik dengan memeriksa detil kalimat pada *tweet*, tidak boleh ada duplikasi atau *tweet* yang isinya sama. Apabila terdapat *tweet* yang sama isinya, maka hanya 1 *tweet* diambil dan yang lainnya dihapus. *Tweet* duplikat ini biasanya dilakukan secara sengaja dengan memanfaatkan robot maupun buzzer, untuk mengcopy *tweet* dan dipost kembali oleh user lainnya. Atau dapat juga dengan melakukan *re-tweet* (RT). Oleh karena itu, untuk mendeteksi *tweet* duplikat, adalah dengan menghapus *hyperlink* dan token "*RT*" (*re-tweet*). Tujuan penghapusan duplikasi adalah agar dalam proses klasifikasi (tahap prediksi) oleh metode *machine learning*, banyaknya *tweet* yang sama tidak akan berkontribusi terhadap penurunan atau kenaikan *score* akurasi secara tidak wajar. Karena hal ini akan berdampak kepada ranking peserta di dalam *leader board* (papan *ranking*). *Dataset* final yang diperoleh, meliputi 1524 data *tweet*.

Dari data ini, diambil sebanyak 300 *tweet* (100 *tweet* untuk masing-masing kelas positif, negatif dan netral), sebagai data *training*. Disediakan sebanyak 2 versi data *training*, yang masing-masing terdiri atas 300 *sample tweet*. Sisanya sebanyak 924 data dipersiapkan sebagai *benchmark* data uji, dan seluruh peserta/peneliti harus melakukan prediksi terhadap data tersebut. Label data uji tidak diberikan kepada peserta. Proses *scoring* evaluasi dilakukan oleh suatu sistem papan *ranking* (*leaderboard*), dan peserta/peneliti mengirim *file* hasil prediksinya ke sistem *leaderboard*.

### 2.1.2 Dataset Open Topic

Selain data *training* yang kecil yang disediakan untuk para peneliti atau peserta *shared task*, kami juga menyediakan data eskternal yang dapat digunakan untuk proses penambahan sampel data. Data eksternal adalah data *tweet* yang telah diberikan label (positif, negatif dan netral) pada isu atau studi kasus yang berbeda.

Data eksternal pertama yang dapat dipakai adalah data *tweet* yang tidak berfokus pada topik atau isu tertentu. Dalam penyebutan nama, data ini diberi nama *Dataset Open Topic*. Data ini diambil dari *twitter* pada rentang bulan Januari 2021 – Februari 2021, tanpa menggunakan kata kunci yang spesifik. Dari proses *scrapping* yang dilakukan didapatkan lebih dari 30 ribu sampel *tweet*. Kemudian dilakukan proses pemberian label oleh 42 orang *anotator*, dengan masing-masingnya mendapatkan tugas sebanyak 2000 *tweet*. Pada tahap pertama ini, diambil 28,000 data sampel untuk diberikan label.

Disebabkan karena tidak adanya topik khusus, banyak *tweet* yang tidak relevan yang diperoleh dari proses *scrapping*, misalnya *tweet* tentang penyebaran pornografi, *tweet* tentang pemasaran produk, *tweet* dengan bahasa asing dan sebagainya. Oleh karena itu, dalam pemberian anotasi, kami membuat kelas tambahan selain positif, negatif dan netral, yaitu kelas *adult* (konten dewasa) dan asing (*tweet* berbahasa asing). Proses pemberian label juga dilakukan secara *majority voting* sebagaimana *Dataset Kaesang*, kecuali untuk *tweet* dengan kelas *adult* dan asing, walaupun dipilih oleh satu orang *anotator* saja, maka *tweet* tersebut dihapus dari dataset. Contoh ilustrasi penilaian label akhir untuk *tweet* dari koleksi open topic dapat dilihat pada Tabel 3 berikut ini.

Tabel 3. Ilustrasi penentuan label pada *dataset open topic*

| No | Kalimat *tweet* | Anotatator | | | | Voted label |
|---|---|---|---|---|---|---|
| | | A1 | A2 | A3 | A4 | |
| 1 | Ready hari ini kediri yang butuh brondong Khusus wanita pasutri dm WA 0881036100224 #gigolokediri #pijatkediri #bokediri #PijatPanggilan #PijatEnak #PijatKediri #PijatWanita #PijatTrenggalek #PijatTulungagung #PijatBlitar #availkediri #bokediri #bisyarsurabaya #bisyarmalang https://t.co/BQBWTFoYvu | **adult** | netral | negatif | **adult** | [dihapus] |
| 2 | @mas__piyuuu Kami menuntut @TMCPoldaMetro agar diproses sesuai hukum yg berlaku di NKRI. Itu pun kalau masih berlaku. @DivHumas_Polri | positif | **netral** | negatif | **netral** | **netral** |
| 3 | SUPER SATURDAY! 😵 Look what we have for you today 😄 We are here till 3pm... @ Central Park, East Ham https://t.co/TskBE7Cbwi | **asing** | **asing** | netral | **asing** | [dihapus] |
| 4 | @samurai264 @Boankrsn @Johny_Ardy @JimlyAs Mengkritik itu gk gitu, ada normanya | netral | netral | **negatif** | **negatif** | [dihapus] |

Proses *detailing* juga dilakukan untuk *dataset open topic* yaitu menghapus semua *tweet* bahasa asing dan konten dewasa, menghapus semua label yang tidak mencapai suara terbanyak, dan menghapus *tweet* yang sama (duplikasi). Proses *detailing* ini menghasilkan *fine grained data tweet* dengan label positif, negatif dan netral yang cocok untuk tugas klasifikasi sentimen sebanyak 7596 sampel *tweet* yang dapat digunakan sebagai data *training* tambahan.

### 2.1.3 Dataset Sentimen Program Vaksinasi Covid-19

Dataset ini disediakan untuk alternatif pilihan bagi para peserta atau peneliti, untuk proses penambahan data *training* (*data agregation*). Dataset ini berasal dari penelitian bersama (*shared task*) di bidang klasifikasi sentimen, dengan fokus permasalahan pada kelas data yang tidak berimbang (*imbalanced data*) [17]. Proses pengembangan dataset dan labeling dilakukan secara *crowdsourcing* dengan setiap *tweet* dinilai oleh 3 orang *anotator* secara suara terbanyak. Jumlah data yang dapat dipakai untuk training adalah 8000 sample, dengan proporsional kelas yang sangat tidak seimbang, seperti diuraikan pada Tabel 1.

Selain dataset yang telah disediakan, peserta *shared task* ini dipersilakan menggunakan dataset lainnya dari penelitian sebelumnya, atau peneliti lain, dengan tidak lupa menyebutkan sumbernya dalam publikasi.

### 2.2. Inter-Annotator Agreement (Persetujuan Antar *Anotator*)

Dua atau lebih *anotator* (penanda, pengklasifikasi, atau pemberi label data) dapat memberikan label berbeda terhadap data yang sama. Dan kebanyakan *anotator* bekerja berdasarkan subjektifitas pribadi terhadap suatu isu atau kasus yang diberikan. Walau bagaimanapun, keberpihakan, khususnya dalam hal klasifikasi perasaan (sentimen) selalu dapat menghinggapi *anotator* dalam melakukan tugasnya memberi label pada kalimat-kalimat yang berisi sentimen.

Oleh karena itu, diperlukan suatu cara untuk mengukur apakah beberapa orang *anotator* yang ditugaskan, telah melakukan proses pelabelan data yang konsisten terhadap sekumpulan data yang sama. *Inter-Annotator Agreement* (IAA) sangat penting dalam bidang yang memerlukan anotasi secara manual oleh manusia [18], [19], walaupun mereka adalah tenaga ahli di bidangnya, seperti komputasi linguistik, analisis konten bahasa alami (*natural language*), penelitian sosial, penelitian di bidang hukum, kedokteran, dan sebagainya. Tujuan utama dari IAA adalah untuk memastikan bahwa anotasi yang diberikan oleh *anotator* adalah handal dan tidak terlalu dipengaruhi oleh subjektivitas individu [20].

Ada beberapa metode yang digunakan untuk mengukur IAA, tergantung pada jenis data dan anotasi. Metrik pengukuran yang umum adalah Cohen's Kappa (κ) yang digunakan untuk dua *anotator*. Ini memperhitungkan kemungkinan kesepakatan yang terjadi secara kebetulan [21]. Namun apabila *anotator* yang ada lebih dari 2 orang, maka digunakan Fleiss' Kappa, yang menghitung rata-rata dari kesepakatan antar *anotator* [20].

Cohen's Kappa diukur dengan menghitung proporsi kesepakatan yang diamati antara *anotator* (*Po*), dan proporsi kesepakatan yang diharapkan secara kebetulan (*Pe*), yang dihitung menurut persamaan (1). Namun metode ini hanya dapat digunakan untuk menghitung kesepakatan di antara 2 *anotator* terhadap 2 label data saja. Misalnya dari 1000 kalimat yang berisi sentimen, hanya diberikan label positif dan negatif saja, dengan melibatkan 2 orang *anotator* saja. Persetujuan antar *anotator* dapat diukur dengan menggunakan persamaan ini.

$$\kappa = \frac{Po - Pe}{1 - Pe} \tag{1}$$

Sedangkan bila jumlah label lebih dari 3 (misalnya positif, negatif dan netral), dan data tersebut diberikan label oleh lebih dari 3 orang, yang masing-masing menandai seluruh kalimat (1000 sampel), maka pengukuran kesepakatan *anotator* dapat diukur menggunakan persamaan Fleiss' Kappa, yaitu rata-rata kesepakatan di antara *anotator*. Pengukuran Fleiss' Kappa dihitung dengan persamaan (2), dengan $\bar{P}$ adalah rata-rata kesepakatan yang diamati, dan $\overline{Pe}$ adalah rata-rata kesepakatan yang diharapkan secara kebetulan.

$$\kappa = \frac{\bar{P} - \overline{Pe}}{1 - \overline{Pe}} \tag{2}$$

**Pengukuran IAA**

Untuk dataset yang dianotasi secara crowdsourcing, sebenarnya pengukuran IAA dengan metode Kappa sebagaimana persamaan 1 dan 2 kurang tepat. Hal ini karena para *anotator* tidak melakukan labeling terhadap keseluruhan data sampel, tetapi hanya sebagian kecil saja. Sehingga sangat terbuka peluang beberapa *anotator* yang bertolak-belakang preferensi sentimennya (subjektivitas) menganotasi *tweet* yang sama. Hal ini dapat berkontribusi pada rendahnya *score* Kappa yang diperoleh, baik itu untuk data dengan 2 kelas (Cohen's Kappa) maupun data dengan 3 kelas atau lebih (Fleiss' Kappa). Sebaliknya, bila kebetulan beberapa *anotator* yang memiliki persamaan preferensi sentimen melakukan anotasi pada sekelompok data yang sama, maka kemungkinan kesepakatan di antara para *anotator* akan lebih tinggi, sehingga *score* Kappa menjadi lebih tinggi pula.

Pengukuran Kappa sejauh penelusuran kami dari berbagai makalah ilmiah, belum ditemukan pada crowdsourcing dengan banyak *anotator*. Pengukuran Kappa dilaporkan bila beberapa orang *anotator* terlibat dalam pelabelan seluruh data, bukan sebagian data saja. Dengan demikian, penentuan label secara *majority voting* sudah cukup untuk finalisasi label data dalam konstruksi dataset penelitian ini.

Namun, apabila tetap diukur IAA menggunakan persamaan Fleiss' Kappa (untuk jumlah *anotator* dan jumlah kelas yang lebih dari 2), maka beberapa orang *anotator* secara kolektif dianggap sebagai single *anotator* (SA). Dalam perhitungan ini, karena setiap *tweet* dianotasi oleh 4 orang, maka terdapat 4 SA kolektif. Selanjutnya pengukuran IAA adalah terhadap SA, bukan terhadap *anotator* secara pribadi. Hal ini berdampak kepada tidak rasionalnya *score* IAA yang didapat, sebagaimana alasan yang telah disampaikan di atas. Tabel 2 menunjukkan hasil perhitungan persetujuan antara pasangan SA, untuk dihitung rata-rata dari persetujuan yang diamati (*Po*), dan persetujuan yang diharapkan secara kebetulan (*Pe*).

Tabel 2. Persetujuan antar Single *Anotator* (kolektif)

| Pasangan SA | *Po* | *Pe* | Cohen's Kappa (antara 2 SA) |
|---|---|---|---|
| SA#1 – SA#2 | 0.6050 | 0.3966 | 0.3453 |
| SA#1 – SA#3 | 0.6190 | 0.3831 | 0.3824 |
| SA#1 – SA#4 | 0.6005 | 0.3578 | 0.3779 |
| SA#2 – SA#3 | 0.6385 | 0.3830 | 0.4141 |
| SA#2 – SA#4 | 0.5970 | 0.3536 | 0.3766 |
| SA#3 – SA#4 | 0.6420 | 0.3996 | 0.4037 |
| **Rata-rata** | $\overline{Po}$ | $\overline{Pe}$ | **Fleiss' Kappa** |
| | 0.6170 | 0.3789 | 0.3833 |

Dari tabel perhitungan persetujuan rata-rata (Fleiss' Kappa) di dalam Tabel 2, dapat disimpulkan berdasarkan interpretasi pengukuran persetujuan *anotator* (IAA) berada dalam rentang penilaian *fair*. Hal ini dinyatakan dalam skala penilaian pada Tabel 3 berikut ini [19].

Tabel 3. Interpretasi Hasil Pengukuran Kappa

| **Rentang Nilai Kappa** | **Interpretasi** |
|---|---|
| 0.00 | *No agreement* |
| 0.01 - 0.20 | *No significant agreement* |
| 0.21 - 0.40 | *Fair agreement* |
| 0.41 - 0.60 | *Moderate agreement* |
| 0.61 - 0.80 | *Substantial agreement* |
| 0.81 - 0.99 | *Almost Perfect agreement* |
| 1.00 | *Perfect agreement* |

### 2.3. *Official Score*

Sebagai penentuan metode yang terbaik, maka metrik evaluasi yang digunakan dalam *shared task* ini adalah *F1-score*. Metode ini dipilih karena mampu mengukur kinerja sistem klasifikasi dalam mendeteksi kelas-kelas yang tidak seimbang. Metrik akurasi tidak dapat dipakai sebagai official *score*, karena akurasi hanya menghitung berapa data yang dapat diklasifikasikan benar sesuai kelasnya, tanpa mempertimbangkan apakah suatu label berjumlah lebih dominan dari yang lainnya. Sebagai contoh, suatu dataset yang memiliki data kelas yang sangat tidak seimbang, misalnya antara kelas negatif dan positif, memiliki perbandingan 90:10 untuk masing-masing kelas. Ketika sistem hanya mampu memprediksi kelas negatif saja dari suatu data uji, dan ketika ada data dengan label *gold standard* bernilai positif, maka akurasi sistem tersebut akan menjadi 90%. Suatu *score* yang sangat tinggi, sedangkan bila ditelaah lebih lanjut, sistem tersebut tidak mampu mendeteksi adanya sentimen positif dari data uji.

### 3. HASIL DAN DISKUSI

Sebuah papan *score* berbasis web disediakan untuk para peneliti men-*submit* hasil prediksinya. *Score* (*F1-score*, *Accuracy*, *Precision* dan *Recall*) yang dihitung secara *macro-average* akan langsung ditampilkan sesaat setelah *submit*. Diberikan 3 kali kesempatan kepada peneliti untuk men-*submit* hasil prediksi. Hasil RUN (submisi) terbaik dari para peneliti ditampilkan di dalam *Leader board* yang dapat diakses oleh peneliti lain sebagai pembanding unjuk kerja.

Dalam keadaan awal (inisialisasi *shared task*), hanya terdapat 2 nilai di dalam sistem *leader board*, yaitu metode *Baseline* SVM (nama peneliti/tim adalah "Admin") dan metode *Optimized* SVM (nama tim adalah "Organizer"), sebagaimana Tabel 4.

### 3.1 Metode *Baseline*

Metode *baseline* dalam berbagai penelitian metode klasifikasi teks merujuk pada pendekatan atau model awal yang digunakan sebagai titik acuan untuk membandingkan performa model dari metode yang lebih kompleks. *Baseline* ini memberikan gambaran tentang performa minimum yang diharapkan dan membantu menilai apakah model baru atau metode yang diusulkan peneliti di dalam *shared task* ini, dapat memberikan peningkatan yang signifikan.

Dalam *shared task* ini, kami memberikan dua *score* evaluasi sebagai referensi untuk acuan. Yang pertama adalah *baseline score*, yaitu *score* terendah yang harus dilampaui oleh peneliti lain untuk dapat melaporkan hasil penelitiannya. Metode *baseline* yang diimplementasikan dalam *shared task* ini adalah SVM (*support vector machine*) Classifier (SVC)[2] tanpa menerapkan langkah optimasi apa pun.

Metode SVM dipilih karena menjadi *state-of-the-art* dari metode *machine learning* konvensional. Tantangan yang besar bagi metode *deep learning* untuk dapat menghasilkan model klasifikasi yang memiliki performa terbaik, karena keterbatasan sumber *data training* yang disediakan. Kami memberi kebebasan kepada para peneliti dan peserta *shared task* ini untuk dapat mengembangkan metodenya sendiri dan langkah optimasi untuk menghasilkan model yang terbaik.

---

[2] https://scikit-learn.org/stable/modules/generated/sklearn.svm.SVC.html

Sebagai fitur klasifikasi, digunakan vektor TF.IDF dengan token berbentuk kata (*word 1-gram*), yang dihitung dari data training yang tersedia (300 sampel) menggunakan *library* tfidf_vectorizer[3]. Pemrosesan teks yang diterapkan adalah pemrosesan dasar seperti *text cleaning* (membersihkan teks dari karakter tunggal, angka, dan tanda baca), menghapus *hiperlink*, mengubah mention menjadi token "*@USER*") dan *case folding* (mengubah seluruh teks menjadi huruf kecil).

Hasil klasifikasi terhadap data testing, yang diukur secara *macro-average* untuk *F1-score, accuracy, precision* dan *recall* secara berturut-turut adalah 0.4038, 0.4545, 0.4953 dan 0.4880, seperti dilaporkan pada Tabel 4. Hasil ini tergolong rendah, karena dari segi nilai *F1-score* di bawah 0.5. Sudah dapat dipastikan sistem gagal dalam melakukan prediksi kelas data uji.

Tabel 4. Nilai acuan bagi para peneliti atau peserta *shared task*

| Metode | Optimasi | RUN | F1-*score* | Accuracy | Precision | Recall |
|---|---|---|---|---|---|---|
| Baseline SVM | - | RUN 1 | 0.4038 | 0.4545 | 0.4953 | 0.4880 |
| Optimzed SVM | KS1 + KS2 | RUN 1 | 0.4546 | 0.5081 | 0.5209 | 0.5873 |
| | KS1 + Cov | RUN 2 | **0.5128** | **0.6121** | 0.5289 | 0.5722 |
| | KS1 + KS2+ Cov | RUN 3 | 0.4953 | 0.5688 | **0.5432** | **0.6098** |

### 3.2 Metode *Optimized*

Nilai referensi yang kedua, menjadi *score* evaluasi acuan untuk sistem-sistem yang sudah mengalami peningkatan (*system enhancement*). Sebagai *score baseline* optimal, kami menerapkan metode SVM yang menggunakan fitur TF.IDF, sebagai fitur yang umum di dalam pemrosesan teks. Langkah optimasi yang dilakukan adalah langkah optimasi dasar, yaitu *data agregation*, penambahan data training dengan data eksternal dalam jumlah tertentu, yang ditentukan secara empiris. Nilai referensi kedua ini menjadi acuan bagi suatu sistem ML yang dikembangkan oleh peneliti lain dalam melakukan optimasi metode ML yang digunakannya. Kami mengharapkan para peneliti dapat melampaui *score baseline* optimal ini.

Dari ketiga RUN yang di-*submit* ke sistem *leader board*, hasil yang terbaik menurut *F1-score* dan *accuracy* adalah pada RUN 2. Sedangkan dari sisi *precision* dan *recall*, yang terbaik adalah RUN 3. Berdasarkan tabel papan *score*, maka yang ditampilkan untuk pemeringkatan peserta *shared task* adalah RUN 2 saja. Bila dibandingkan dengan baseline SVM tanpa optimasi, hasil optimasi dengan penambahan data Kaesang saja (KS1 dan KS2) dapat meningkatkan performa sebesar *F1-score* 5.08%. Sedangkan dengan menambahkan dataset Covid pada RUN 3, performa mengalami peningkatan sebesar 9.15%. Namun sedikit anomali terdapat pada RUN 2, di mana pada saat dataset KS2 tidak digunakan, performa justru paling baik, yaitu mencapai 51.28%, selisih 1,75% dari RUN 3, dan 10,90% dari baseline tanpa optimasi.

Terlihat jelas bahwa penambahan dataset *tweet*, walaupun berbeda isu yang dibahas, dapat meningkatkan hasil klasifikasi dengan cukup efektif. Hasil ini akan dibandingkan nantinya dengan hasil-hasil yang dicapai oleh para peneliti lain. Sampai saat ini *shared task* ini telah menarik lebih dari 16 orang peneliti untuk mengembangkan metode klasifikasi masing-masing.

### 4. KESIMPULAN

Penelitian ini mengangkat permasalahan keterbatasan dataset yang dapat digunakan untuk pelatihan model machine learning pada tugas klasifikasi teks. Kontribusi dari penelitian ini adalah dataset yang dapat dijadikan benchmark penelitian yang dapat digunakan oleh para peneliti untuk menguji metode yang dikembangkan. Penelitian ini dikembangkan sebagai shared task, sehingga mengundang banyak peneliti untuk berpartisipasi dalam tugas klasifikasi dengan tantangan yang diberikan. Penelitian ini juga menunjukkan langkah optimasi dari segi dataset, yaitu dengan menambahkan data training dari topik yang berbeda (external dataset) dapat meningkatkan performa klasifikasi secara signifikan, yaitu 10.9%. Beberapa langkah optimasi lainnya dibahas dalam makalah ini, dapat diujicoba oleh para peserta shared task untuk dikembangkan, dan model paling optimal yang berhasil diperoleh dapat diujikan pada data testing yang tersedia.

**ACKNOWLEDGMENTS**

Apresiasi kami sampaikan kepada para mahasiswa Teknik Informatika UIN Suska Riau, yang sudah membantu dalam hal pembangunan dataset, yang terlibat mulai dari proses *scrapping* data dari *Twitter*, sampai kepada tim *anotator* yang sudah bekerja memberi label pada data. Apresiasi juga kami sampaikan kepada para mahasiswa yang sudah berpartisipasi dalam *shared task* ini dan mengerahkan *effort* yang besar untuk mengembangkan model klasifikasi dengan berbagai metode.

---

[3] https://scikit-learn.org/stable/modules/generated/sklearn.feature_extraction.text.TfidfVectorizer.html

**REFERENCES**

[1] S. A. Salahudeen *et al.*, "HausaNLP at SemEval-2023 Task 12: Leveraging African Low Resource *Tweet*Data for Sentiment Analysis," in *Proceedings of the The 17th International Workshop on Semantic Evaluation (SemEval-2023)*, Stroudsburg, PA, USA: Association for Computational Linguistics, 2023. doi: 10.18653/v1/2023.semeval-1.6.

[2] R. Vindua and A. U. Zailani, "Analisis Sentimen Pemilu Indonesia Tahun 2024 Dari Media Sosial *Twitter* Menggunakan Python," *JURIKOM (Jurnal Riset Komputer)*, vol. 10, no. 2, Apr. 2023, doi: 10.30865/jurikom.v10i2.5945.

[3] I. H. Hasibuan, E. Budianita, and S. Agustian, "Klasifikasi Sentimen Komentar Youtube Tentang Pembatalan Indonesia Sebagai Tuan Rumah Piala Dunia U-20 Menggunakan Algoritma Naïve Bayes Classifer," *Jurnal Sistem Komputer dan Informatika (JSON)*, vol. 5, no. 2, pp. 249–257, 2023, doi: 10.30865/json.v5i2.7096.

[4] A. Naldi and S. Agustian, "Klasifikasi sentimen Vaksin Covid-19 menggunakan K-Nearest…," *ZONAsi: Jurnal Sistem Informasi*, vol. 5, no. 2, pp. 323–333, 2023.

[5] S. Azhar, M. Fikry, S. Agustian, and I. Afrianty, "Klasifikasi Sentimen Masyarakat di *Twitter* Terhadap Ganjar Pranowo dengan Metode Support Vector Machine," *KLIK: Kajian Ilmiah Informatika dan Komputer* , vol. 4, no. 3, pp. 1660–1667, 2023, doi: 10.30865/klik.v4i3.1537.

[6] E. H. Muktafin, K. Kusrini, and E. T. Luthfi, "Analisis Sentimen pada Ulasan Pembelian Produk di Marketplace Shopee Menggunakan Pendekatan Natural Language Processing," *Jurnal Eksplora Informatika*, vol. 10, no. 1, pp. 32–42, Sep. 2020, doi: 10.30864/eksplora.v10i1.390.

[7] S. Satapara *et al.*, "Hate-Speech Identification in Sinhala and Gujarati under Creative Commons License Attribution 4.0 International (CC BY 4.0).," in *Overview of the HASOC Subtrack at FIRE 2023*, 2023. [Online]. Available: http://ceur-ws.org

[8] J. Devlin, M.-W. Chang, K. Lee, and K. Toutanova, "BERT: Pre-training of Deep Bidirectional Transformers for Language Understanding," *ArXiv*. 2019.

[9] T. B. Brown *et al.*, "Language Models are Few-Shot Learners," *ArXiv*. 2020.

[10] M. Wankhade, A. C. S. Rao, and C. Kulkarni, "A survey on sentiment analysis methods, applications, and challenges," *Artif Intell Rev*, vol. 55, no. 7, pp. 5731–5780, Oct. 2022, doi: 10.1007/s10462-022-10144-1.

[11] S. Agustian and H. Takamura, "UINSUSKA-TiTech at SemEval-2017 Task 3: Exploiting Word Importance Levels for Similarity Features for CQA," in *Proceedings of the 11th International Workshop on Semantic Evaluation  (SemEval-2017)*, Stroudsburg, PA, USA: Association for Computational Linguistics, 2017, pp. 370–374. doi: 10.18653/v1/S17-2061.

[12] F. Ihsan, I. Iskandar, N. S. Harahap, and S. Agustian, "Decision tree algorithm for multi-label hate speech and abusive language detection in Indonesian *Twitter*," *Jurnal Teknologi dan Sistem Komputer*, vol. 9, no. 4, pp. 199–204, Oct. 2021, doi: 10.14710/jtsiskom.2021.13907.

[13] T. Yu and H. Zhu, "Hyper-Parameter Optimization: A Review of Algorithms and Applications," *ArXiv*, vol. abs/2003.05689, 2020, [Online]. Available: https://api.semanticscholar.org/CorpusID:212675087

[14] Y. Roh, G. Heo, and S. E. Whang, "A Survey on Data Collection for Machine Learning: a Big Data -- AI Integration Perspective," *ArXiv*, Nov. 2018.

[15] V. Batanović, M. Cvetanović, and B. Nikolić, "A versatile framework for resource-limited sentiment articulation, annotation, and analysis of short texts," *PLoS One*, vol. 15, no. 11, Nov. 2020, doi: 10.1371/journal.pone.0242050.

[16] M. Nabil, M. Aly, and A. Atiya, "ASTD: Arabic Sentiment *Tweet*s Dataset," in *Proceedings of the 2015 Conference on Empirical Methods in Natural Language Processing*, Stroudsburg, PA, USA: Association for Computational Linguistics, 2015. doi: 10.18653/v1/D15-1299.

[17] P. Yohana, S. Agustian, and S. Kurnia Gusti, "Klasifikasi Sentimen Masyarakat terhadap Kebijakan Vaksin Covid-19 pada *Twitter* dengan Imbalance Classes Menggunakan Naive Bayes," in *SNTIKI (Seminar Nasional Teknologi Informasi, Komunikasi dan Industri)*, 2022. Accessed: Jun. 15, 2024. [Online]. Available: https://ejournal.uin-suska.ac.id/index.php/SNTIKI/article/download/19012/8336

[18] R. Artstein and M. Poesio, "Inter-Coder Agreement for Computational Linguistics," *Computational Linguistics*, vol. 34, no. 4, pp. 555–596, Dec. 2008, doi: 10.1162/coli.07-034-R2.

[19] J. R. Landis and G. G. Koch, "The Measurement of Observer Agreement for Categorical Data," *Biometrics*, vol. 33, no. 1, p. 159, Mar. 1977, doi: 10.2307/2529310.

[20] K. Krippendorff, *Content Analysis: An Introduction to Its Methodology*. Sage Publications, 2004.

[21] J. Cohen, "A Coefficient of Agreement for Nominal Scales," *Educ Psychol Meas*, vol. 20, no. 1, pp. 37–46, Apr. 1960, doi: 10.1177/001316446002000104.